\documentclass{article}

\PassOptionsToPackage{numbers, sort&compress}{natbib}
 \usepackage[final]{neurips_2025_ml4ps}

\usepackage[utf8]{inputenc} 
\usepackage[T1]{fontenc}    
\usepackage{hyperref}       
\usepackage{url}            
\usepackage{booktabs}       
\usepackage{amsfonts}       
\usepackage{nicefrac}       
\usepackage{microtype}      
\usepackage{xcolor}         
\usepackage{graphicx}
\usepackage{amsmath}
\usepackage{amssymb}
\usepackage{subcaption}

\newcommand{\dd}{\mathop{}\!\mathrm{d}}

\title{Lagrangian neural ODEs: Measuring the existence of a Lagrangian with Helmholtz metrics}

\author{%
  Luca Wolf\\
  Interdisciplinary Center for Scientific Computing,\\
  Heidelberg University,\\
  Im Neuenheimer Feld 205, 69120 Heidelberg, Germany\\
  \texttt{luca.wolf@stud.uni-heidelberg.de} \\
    \AND
     Tobias Buck\\
  Interdisciplinary Center for Scientific Computing,\\ Heidelberg University,\\ Im Neuenheimer Feld 205, 69120 Heidelberg, Germany\\
  \texttt{tobias.buck@iwr.uni-heidelberg.de} \\
    \And
     Bj{\"o}rn Malte Sch{\"a}fer\\
  Astronomisches Rechen-Institut der Universit{\"a}t Heidelberg,\\ Philosophenweg 12, 69120 Heidelberg, Germany\\
  \texttt{bjoern.malte.schaefer@uni-heidelberg.de} \\
}

\begin{document}

\maketitle

\begin{abstract}
  Neural ODEs are a widely used, powerful machine learning technique in particular for physics. However, not every solution is physical in that it is an Euler-Lagrange equation. We present Helmholtz metrics to quantify this resemblance for a given ODE and demonstrate their capabilities on several fundamental systems with noise. We combine them with a second order neural ODE to form a Lagrangian neural ODE, which allows to learn Euler-Lagrange equations in a direct fashion and with zero additional inference cost. We demonstrate that, using only positional data, they can distinguish Lagrangian and non-Lagrangian systems and improve the neural ODE solutions.
\end{abstract}

\section{Differential equations in physics and Machine Learning}
Ordinary differential equations (ODEs) are essential for describing real-world dynamical systems in the sciences and beyond. This led to the development of neural ODEs \cite{neuralODE}, which use a neural network (NN) $h_\theta$ to find the ODE $\dot{s} = h_\theta(t,s)$ based on observed data of the system $s(t)$. To achieve this, they numerically integrate $h_\theta$ from given initial conditions to the desired points in time. The result can then be compared to the data by means of a regression loss $\mathcal{L}_R$, like mean squared error (MSE) and the network is trained with standard backpropagation.\par

However, not every ODE is eligible to be a (fundamental) physical law: As one of the most fundamental principles in theoretical physics, the \emph{principle of stationary action} demands any trajectory $(x, \dot{x})$ to be a stationary solution of the action functional
\begin{equation}
    S[x]=\int_{t_A}^{t_B}L(t,x(t),\dot{x}(t))\, \dd t
\end{equation}
for a Lagrangian $L$ specific to the system. This is the case if and only if the ODE, called the \emph{Euler-Lagrange equation},
\begin{equation}\label{eq:EL}
    \mathcal{E}_x[L]:=\frac{\partial L}{\partial x}-\frac{\dd}{\dd t}\frac{\partial L}{\partial\dot{x}}=g\ddot{x} - \left(\frac{\partial^2 L}{\partial x \partial \dot{x}} \dot{x} + \frac{\partial^2 L}{\partial t \partial \dot{x}} - \frac{\partial L}{\partial x}\right)=0,
\end{equation}
holds. If the Hessian with respect to $\dot{x}$, $g(t,x,\dot{x})$, is invertible, it may be written in an explicit form. The theorem of Ostrogradsky further implies that this ODE is always of second order \cite{ostrogradski_1, ostrogradski_2}.\par

To incorporate the higher order dynamics, the neural ODE state can be augmented with arbitrary additional dimensions $s=(x,a)$ \cite{anode} or made explicitly second order,
\begin{equation}\label{eq:second_order_neural_ode}
    \frac{\dd}{\dd t}\begin{pmatrix}x\\v\end{pmatrix}=\begin{pmatrix}v\\f_{\theta_1}(t,x,\dot{x})\end{pmatrix}, \quad v_0=\mathrm{NN}_{\theta_3}(x_0),
\end{equation}
where now the NN $f_{\theta_1}$ directly models $\ddot{x}$, which performs better on physical systems \cite{sonode}. The initial condition for $v\equiv\dot{x}$ can be learned by a NN. Then, only positional data $x(t)$ is required for training.\par

However, not every second order ODE $\ddot{x}=f(t,x,\dot{x})$ fulfils the stationary action principle and originates from a Lagrangian via \eqref{eq:EL}. The \textbf{Helmholtz conditions} \cite{douglas} specify if this is the case for an ODE $f$. This provides the basis for our work: We develop \emph{Helmholtz metrics}, to quantify how close a given ODE resembles an Euler-Lagrange equation. They can then be used as a regularisation for a neural ODE, such that it converges to a true physical law. These \emph{Lagrangian neural ODEs} consequently promise more desirable properties. If a given system does not originate from a Lagrangian, the convergence should fail, indicating that this is the case.\par

Not satisfying the stationary action principle has profound physical implications: It may imply that the system is open, dissipative or has similar inconsistencies, e.g., with its description. This is best illustrated with the damped harmonic oscillator, where no time independent Lagrangian exists. And indeed, the damping is not a fundamental principle, but a manifestation of the (fundamental) microscopic interaction with, e.g., air molecules, which are neglected in the description.

\subsection{Related Work}
Lagrangian neural networks (LNNs) \cite{LNN} accomplish a similar goal, by fulfilling the stationary action principle by construction. A NN directly predicts the Lagrangian and the ODE $f$ is constructed from the Euler-Lagrange equations by automatic differentiation. Our method can essentially be viewed as the inverse approach. While the Euler-Lagrange equations are not exactly fulfilled and the Lagrangian can only be recovered in part, our approach also has some advantages: Training the neural ODE does neither require forward computation of nor backpropagation through the Euler-Lagrange equations. This promises higher stability and lower computational cost. Since the Helmholtz metrics only act as regularisation, there is no added inference cost with respect to usual neural ODEs. Due to the adjustable regularisation strength, Lagrangian neural ODEs also open the possibility of bringing ODEs closer to an Euler-Lagrange equation, leveraging their benefits without completely restricting to such an ODE. Hamiltonian neural networks \cite{HNN} essentially work like LNNs, but they operate in the equivalent formalism of Hamiltonian mechanics instead.

\section{Method: Lagrangian neural ODEs and Helmholtz metrics}
Given a second order ODE $\ddot{x}=f(t,x,\dot{x})$, define
\begin{equation}\label{eq:douglas_phi}
    \Phi = \frac{1}{2}\frac{\dd}{\dd t}\frac{\partial f}{\partial \dot{x}}-\frac{\partial f}{\partial x}-\left(\frac{1}{2}\frac{\partial f}{\partial \dot{x}}\right)^2.
\end{equation}
The ODE then originates from a Lagrangian in the sense of \eqref{eq:EL}, i.e., $g(\ddot{x}-f)=\mathcal{E}_x[L]$ if and only if there exists a non-singular symmetric matrix $g$, such that
\begin{equation}\label{eq:Douglas_Helm}
g \Phi = (g \Phi)^{\top} ,\quad
\frac{\dd g}{\dd t} + \dfrac{1}{2} \left(\frac{\partial f}{\partial \dot{x}}\right)^{\top} g + \dfrac{1}{2} g\frac{\partial f}{\partial \dot{x}} = 0 ,\quad
\frac{\partial g}{\partial \dot{x}} = \left(\frac{\partial g}{\partial \dot{x}}\right)^{\top}.
\end{equation}
If these \emph{Helmholtz conditions} are fulfilled, $g$ is the Hessian of the Lagrangian, see \eqref{eq:EL} \cite{douglas, douglas_long}.\par

The central idea of this work is to add the MSE of the residuals of \eqref{eq:Douglas_Helm}, $\mathcal{R}_i$, as a physics informed loss to the training of \eqref{eq:second_order_neural_ode}. However, this is not entirely straightforward, as the existence of the matrix-valued function $g(t,x,\dot{x})$ must be determined and quantified in a differentiable way. Here, we decide to train a NN $g_{\theta_2}$ to minimize $\mathcal{L}_H=\operatorname{MSE}(\sum_i\mathcal{R}_i)$ and take the final loss value as the measure of satisfaction of \eqref{eq:Douglas_Helm}. The output of the NN is transformed with $\sinh$, to handle possible exponential behaviour in the second condition in \eqref{eq:Douglas_Helm}.\par

The matrix can be made symmetric by construction. Making it invertible by construction usually leads to issues with differentiability and still allows quasi-singular matrices that minimize $\mathcal{R}_i$ just by learning small eigenvalues. For that reason, the $\mathcal{R}_i$ are first normalized by the smallest absolute eigenvalue, or the $l(-2)$-norm of $g$: $\lambda_\text{min}=\|g\|_{-2}=\inf_{\|v\|_2\le1} \|gv\|_2$. That is, the residuals must be small not in terms of absolute size but relative to the smallest value $g$ may map (unit vectors) to. This enforces non-singularity in a natural, scale-aware way. We refer to this construct as Helmholtz metric, which forms the core novelty and primary contribution of this work, providing a differentiable measure of an ODE being an Euler-Lagrange equation.\par
\begin{figure}[htb]
    \centering
    \includegraphics[width=0.9\linewidth]{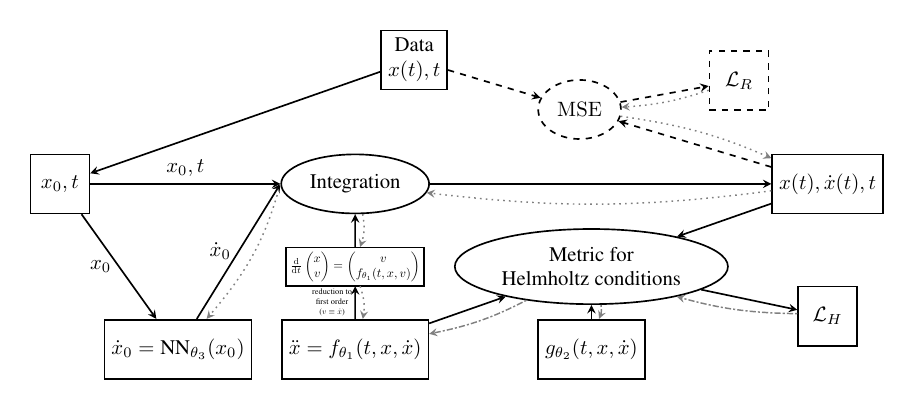}
    
    \caption{Lagrangian neural ODE model. Gradients are represented by gray arrows.}
    \label{fig:flowchart}
\end{figure}

These Helmholtz metrics can then be combined with the neural ODE to give a Lagrangian neural ODE, depicted in Fig.~\ref{fig:flowchart}. 
Fortunately, it is not necessary to retrain $g_{\theta_2}$ for every step and a joint minimization of $\mathcal{L}_\text{tot}=\mathcal{L}_R+\mathcal{L_H}$ is possible. For this multi objective optimization, see e.g. \cite{moo_mtl, paretoMTL}, to succeed we employ several strategies: We clip $\|\nabla_{\theta_{1}}\mathcal{L}_H\|$ to $c_1\approx0.05$ and combine the progressive inclusion of time steps during training, a standard method to avoid local minima \cite{kidgerneuralODEs}, with a high initial learning rate. This assures the model is data-dominated from the start and does not converge, e.g., to the Euler-Lagrange equation of an entirely different system. Note also, that $\mathcal{L}_H$ is evaluated at the predicted trajectories, but not differentiated with respect to them during backpropagation.

\section{Application and evaluation}
To demonstrate both Helmholtz metrics and Lagrangian neural ODEs, we train on (data of) well-understood toy systems. As is widely established, the Softplus activation function is used for the neural ODE $f_{\theta_1}$ \cite{torchdiffeq, LNN}, which also ensures smooth derivatives in \eqref{eq:Douglas_Helm}. The same applies for $g_{\theta_2}$. We use networks with 1 layer of 16 neurons, 2 layers of 64 neurons and 3 layers of 16 neurons for $f_{\theta_1}$, $g_{\theta_2}$ and $\mathrm{NN}_{\theta_3}$, respectively. Optimization is done with RAdam \cite{RAdam} and a batch size of 128, the learning rate (lr) $10^{-1}\ge\text{lr}_0\ge10^{-4}$ reduces on plateau. We use Python, PyTorch and torchdiffeq \cite{torchdiffeq} for our implementation, the code is available on GitHub\footnote{\url{https://github.com/luwo9/LagrangianNeuralODEs}}, including documentation and used scripts.
\subsection{Helmholtz metrics on analytical ODEs}
\begin{figure}[h]

\begin{subfigure}{0.33\linewidth}
    \includegraphics[width=\linewidth]{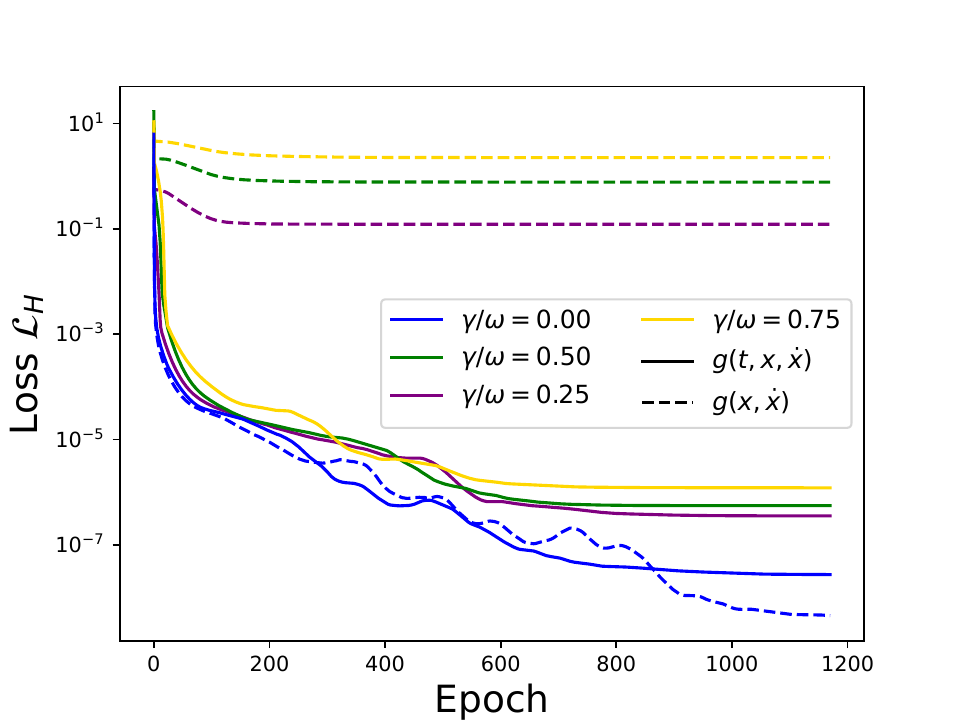}
\end{subfigure}
\begin{subfigure}{0.33\linewidth}
    \includegraphics[width=\linewidth]{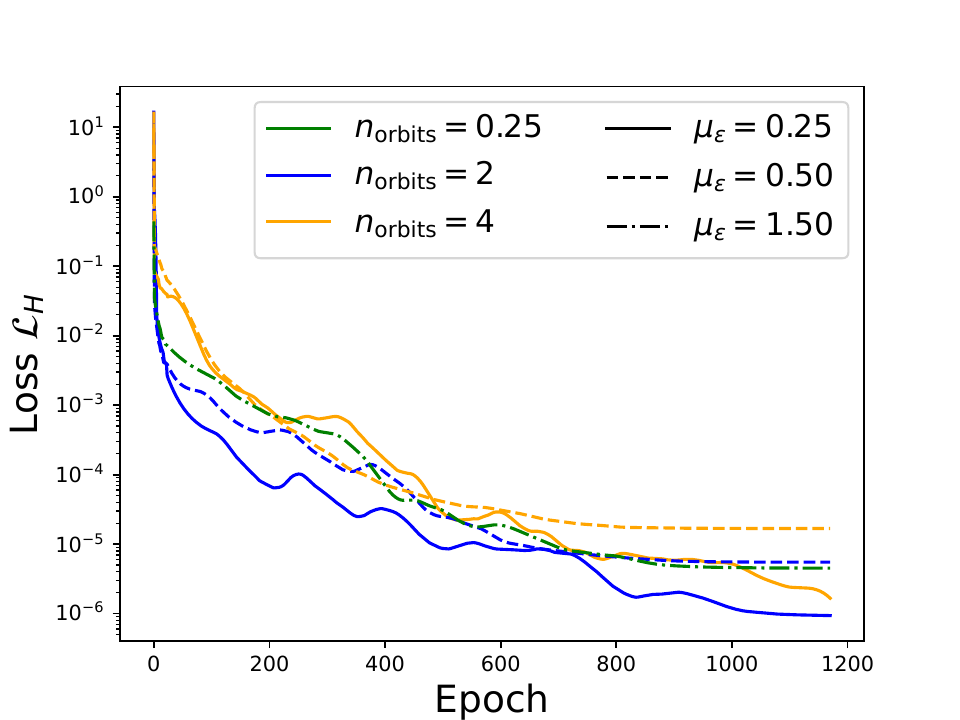}
\end{subfigure}
\begin{subfigure}{0.33\linewidth}
    \includegraphics[width=\linewidth]{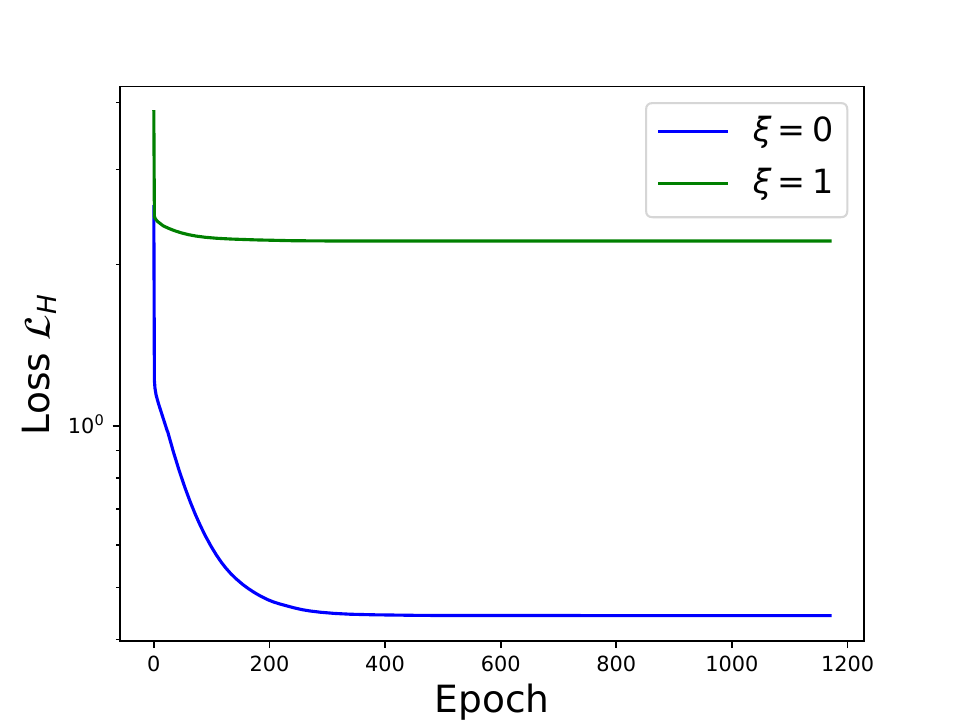}
\end{subfigure}
    \caption{Smoothed loss curves for training Helmholtz metrics on different ODEs for different settings. Left: Oscillator with and without time dependence of $g$. Middle: Kepler problem with different mean eccentricity $\mu_\varepsilon$. Right: non-Lagrangian ODEs. Vertical axes are scaled differently.}
    \label{figlosses_analyt}

\end{figure}

As a first test, we train the Helmholtz metric on fixed, analytical ODEs supplying both $x$ and $\dot{x}$: The 2D damped oscillator, the Kepler problem and two non-Lagrangian ODEs \cite{douglas, douglas_long}
\begin{equation}
    \begin{pmatrix}\ddot{x}_1\\\ddot{x}_2\end{pmatrix} =-\Omega \begin{pmatrix}x_1\\x_2\end{pmatrix} - \Gamma \begin{pmatrix}\dot{x}_1\\\dot{x}_2\end{pmatrix}, \quad
    \begin{pmatrix}
        \ddot{r} \\
        \ddot{\varphi}
    \end{pmatrix}
    =
    \begin{pmatrix}
        r\dot{\varphi}^2 - \frac{G M}{r^2} \\
        -\frac{2 \dot{r} \dot{\varphi}}{r}
    \end{pmatrix}, \quad
    \begin{pmatrix}
    \ddot{x}_1 \\
    \ddot{x}_2
    \end{pmatrix}
    =
    \begin{pmatrix}
    x_1^2 + x_2^2 \\
    \xi x_1
    \end{pmatrix},
\end{equation}

with spring matrix $\Omega=\operatorname{diag}(\omega_1^2, \omega^2_2)$ and damping matrix $\Gamma=\operatorname{diag}(\gamma_1, \gamma_2)$ and where $\xi\in\{1,0\}$. The Kepler problem and oscillator originate from a Lagrangian, though the explicit time dependence of both $L$ and $g$ is required if damping occurs ($\gamma\neq0$). Both $\omega$ and $GM$ can be parametrized as $n_\text{periods}$. For the Kepler problem, the initial conditions are parametrized by semi-latus rectum $p$ and eccentricity $\varepsilon$ of the conic section orbit. For the oscillator, $n_\text{periods}=3$. After training on 6000 trajectories with artificial 5\% noise and initial conditions sampled from Gaussians we obtain the loss curves depicted in Fig.~\ref{figlosses_analyt}.\par

The final, absolute size of $\mathcal{L}_H$ is not of significance, when comparing different systems, due to the direct dependence on the magnitude of $f$ in \eqref{eq:Douglas_Helm}. However, based on the loss improvement, one can readily verify that a solution is found for the Kepler problem and for the damped oscillator if undamped, $\gamma=0$, or time dependent $g=g(t,x,\dot{x})$. Similarly, for the other systems, no Lagrangian exists and the Helmholtz metric finds no solution, i.e, $\mathcal{L}_H$ improves only minimally. To confirm that the learnt solution is actually physical, we compare the predicted $g$ with the analytic solution, the Hessian of the Lagrangian. We take the absolute error for components of $g$ that are expected to be 0 and the relative error for all other elements. Collecting over different parameters, trajectories, time steps and matrix components we obtain median (20th–80th percentile): $3.7\times 10^{-4} \; (1.1\times 10^{-4} \;-\; 1.6\times 10^{-3})$ for the Kepler problem on a test set, demonstrating a close resemblance.\par

\subsection{Lagrangian neural ODEs on data}
\begin{figure}[h]

\begin{subfigure}[t]{0.64\linewidth}
\begin{subfigure}{0.49\linewidth}
    \includegraphics[width=\linewidth]{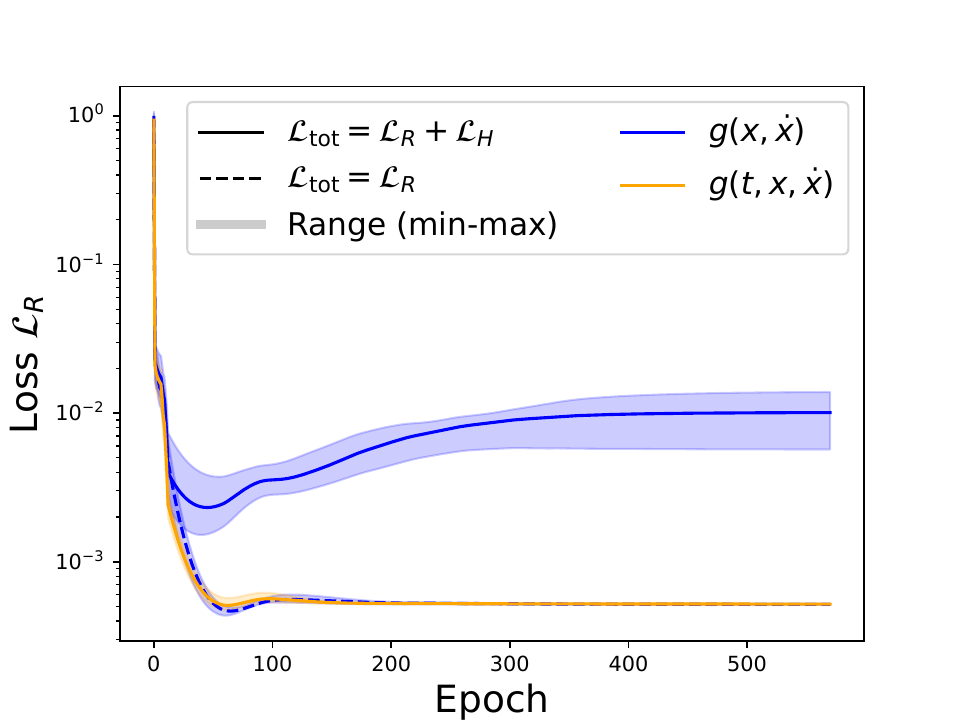}
\end{subfigure}
\begin{subfigure}{0.49\linewidth}
    \includegraphics[width=\linewidth]{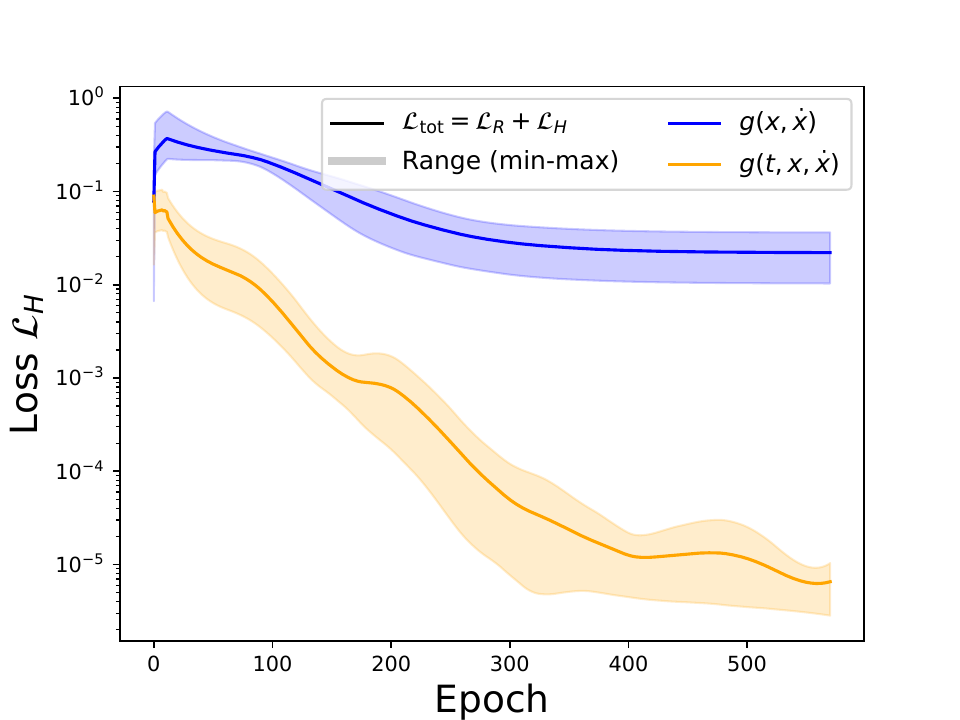}
\end{subfigure}
\caption{Smoothed loss curves for Lagrangian neural ODEs on the damped oscillator. Left: Regression loss $\mathcal{L}_R$. Right: Helmholtz metric $\mathcal{L}_H$.}
\label{fig:losses_neural}
\end{subfigure}
\hfill
\begin{subfigure}[t]{0.32\linewidth}
    \includegraphics[width=0.98\linewidth]{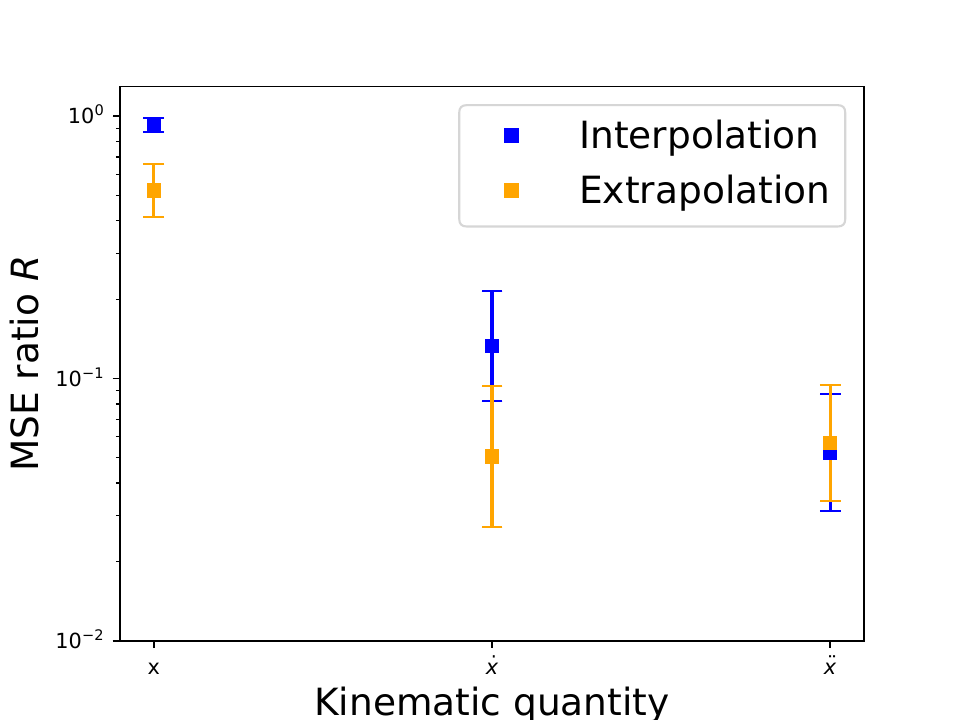}
    \caption{MSE ratio $R$, of regularized and unregularized ODE with 95\% CI.}
    \label{fig:MSE_impr}
\end{subfigure}
    \caption{Results of training a Lagrangian neural ODE on an oscillator.}

\end{figure}

Having established reliance on our newly derived Helmholtz metrics, we now train the full Lagrangian neural ODEs only on positional data. We again use the harmonic oscillator as it allows in one system to investigate: (i) The un-regularized case, as a baseline for neural ODEs with $f(x,\dot{x})$, (ii) the regularized, time independent case with $f(x,\dot{x})$ and $g(x,\dot{x})$, where no Lagrangian exists and (iii) the regularized, time dependent case with $f(t,x,\dot{x})$ and $g(t,x,\dot{x})$, where a Lagrangian exists. With $n_\text{periods}=3$ and $\gamma=0.5\omega$ we train on 6000 5\%-noise trajectories, where $x_0$ is sampled from a Gaussian $\mathcal{N}(\mu=1,\sigma=1)$ and $\dot{x}_0=|x_0|$, to ensure correlation, see \eqref{eq:second_order_neural_ode}. The trajectories are supplied at $n_t=30$ equidistant time points. The mean loss curves and their fluctuation after training 5 models each with $\text{lr}_0=0.07$ are shown in Fig.~\ref{fig:losses_neural}.\par

Evidently, $\mathcal{L}_R$ converges equally well and to the same value for baseline (i) (dashed, blue) and the case with Lagrangian (iii) (solid, orange), indicating that Lagrangian neural ODEs provide no disadvantage here. In contrast, the case without Lagrangian (ii) (solid, blue), performs significantly worse in terms of $\mathcal{L}_R$. This should allow to clearly distinguish whether a given system does or does not originate from a Lagrangian, but a rigorous evaluation on different systems remains part of future work. Further, $\mathcal{L}_H$ decreases only little for the case without Lagrangian (ii) (solid, blue) (likely, as the ODE is distorted and $\mathcal{L}_R$ worsens), but significantly for the case with Lagrangian (iii) (solid, orange).\par

While we found several advantages of Lagrangian neural ODEs, if a Lagrangian exists, a comprehensive quantitative investigation and a comparison to, say LNNs, is still in progress and will be reported in future work. However, we give a first result here: We train 40 models each regularized and unregularized on the oscillator, with $\gamma_{1,2}=0$, $n_t=7$, $n_\text{periods}=1$, $x_0\sim\mathcal{N}(1,0.1)$ and $\dot{x}_0=0.1|x_0|$, as detailed above and record the MSE when comparing to a high resolution ground truth for $x$, $\dot{x}$ and $\ddot{x}$, including extrapolation up two twice the training time interval. We then perform Welch's $t$-test on $l=\log\operatorname{MSE}$ and report the results in terms of MSE ratio $R=\exp(l_\text{regularized}-l_\text{unregularized})$ in Fig.~\ref{fig:MSE_impr}. We achieve a significant improvement in all cases, in particular, for $\dot{x}$ and $\ddot{x}$ and extrapolation.

\section{Conclusion and outlook}
We constructed Helmholtz metrics that, for a given ODE, can quantify their resemblance to an Euler-Lagrange equation. This was demonstrated for several fundamental systems with noise. They also correctly reconstruct the Hessian of the Lagrangian with high accuracy. These metrics can be used as regularisation for second order neural ODEs, to converge to an Euler-Lagrange equation. These allow to determine, only from positional data, whether a Lagrangian exists by their convergence behaviour. If it exists, the solution has preferable properties, as was shown by first results of ground truth MSE. However, a more comprehensive analysis, as well as comparison to some related work, is part of future work.

\begin{ack}
This work is funded by the Carl-Zeiss-Stiftung through the NEXUS program. This work was supported by the Deutsche Forschungsgemeinschaft (DFG, German Research Foundation) under Germany’s Excellence Strategy EXC 2181/1 - 390900948 (the Heidelberg STRUCTURES Ex-
cellence Cluster). We acknowledge the usage of the AI-clusters Tom and Jerry funded by the Field of Focus 2 of Heidelberg University.
\end{ack}

\bibliography{bibliography}

\end{document}